# Dynamic Importance Learning using Fisher Information Matrix (FIM) for Nonlinear Dynamic Mapping

**Vahid MohammadZadeh Eivaghi**

Department of Electrical Engineering K. N. Toosi University of Technology Tehran, Iran, email: vmohammadzadeh@email.kntu.ac.ir

**Mahdi Aliyari Shoorehdeli**

Department of Electrical Engineering K. N. Toosi University of Technology Tehran, Iran, email: aliyari@kntu.ac.ir

## Abstract

Understanding output variance is critical in modeling nonlinear dynamic systems, as it reflects the system's sensitivity to input variations and feature interactions. This work presents a methodology for dynamically determining relevance scores in black-box models while ensuring interpretability through an embedded decision module. This interpretable module, integrated into the first layer of the model, employs the Fisher Information Matrix (FIM) and logistic regression to compute relevance scores, interpreted as the probabilities of input neurons being active based on their contribution to the output variance. The proposed method leverages a gradient-based framework to uncover the importance of variance-driven features, capturing both individual contributions and complex feature interactions. These relevance scores are applied through element-wise transformations of the inputs, enabling the black-box model to prioritize features dynamically based on their impact on system output. This approach effectively bridges interpretability with the intricate modeling of nonlinear dynamics and time-dependent interactions. Simulation results demonstrate the method's ability to infer feature interactions while achieving superior performance in feature relevance compared to existing techniques. The practical utility of this approach is showcased through its application to an industrial pH neutralization process, where critical system dynamics are uncovered.

## 1. Introduction

The problem of nonlinear system identification deals with finding a mathematical description of the dynamic behavior of a system from measured data $\{x_t, y_t\}_{t=1}^N$ to provide accurate prediction of the future behavior given input(s) [1] and [2]:

$$y_t = f_\theta(x_t) + e_t, \qquad E(e_t) = 0, E(e_t e_t) = R \tag{1}$$

Where the $y_t$ is the system output(s), $e_t$ is iid noise sequence of zero mean and finite variance, and input(s) $x_t$ may be a lagged window of system input(s), output(s), and error in general, i.e. $x_t = [y_{t-k_1}, \dots y_{t-1}, u_{t-k_2}, \dots, u_{t-1}, e_{t-k_3}, \dots, e_{t-1}]$. Relationship (1) represents a large class of nonlinear models. The well-known nonlinear autoregressive model with exogenous input (NARX) can be easily instantiated from (1) by skipping those terms related to the error signal $e_t$. Regardless of the model one adopts, the identification problem is usually done in two steps: identification of influential variables and nonlinear function approximation [3]. The former is the topic of variable selection within statistical learning theory. The latter, however, is a mature topic in system identification. A popular approach for realizing function approximation in the literature is to express the unknown function $f$ as the linear combination of some nonlinear functions called basis function [1].

$$y_t = \sum_{k=1}^{M} \alpha_k \phi_k(x_t; \theta) + e_t \tag{2}$$

There is no limitation on how the basis functions are modeled. They can be modeled by black box models like a multi-layered perceptron, Gaussian process, polynomial approximation, functional expansion, and deep learning or transparent models like lookup tables, locally linear models, and fuzzy approximation [1]. For many real-world problems, contributing variables included in $x_t$ are sparse, so not all variables should be included in modeling [4]. It allows us to understand the impact of input variables, rank them, select the most influential inputs, and drop non-relevant ones from the model. This is remarkably important for improving the model's accuracy, revealing the complex interaction among variables in a nonlinear context, and improving the model's interpretability in applications like condition monitoring and fault diagnosis. In this regard, the most important part of an identification problem is inferring the influential variables. This is the topic of model order determination in system identification communities [1]. Focusing on the black-box basis function in relationship (2), in this study, we will address the problem of feature importance determination of nonlinear dynamic systems in an end-to-end manner. Despite the scattered attention received from the control communities to embed the important dynamics into the modeling process, feature selection techniques have been addressed in data-driven modeling and machine-learning communities [5], [6], [7], and [8]. Regarding predictive modeling as a general concept common in control theory and machine learning, the variable importance determination methods can be used to score the importance dynamic in nonlinear dynamic modeling with special attention on the temporal nature of input/output data utilized for identifying dynamic systems.

Feature selection is crucial in building data-driven predictive modeling, focusing on identifying and retaining the most relevant features from a dataset. The goal is to improve the model's performance by eliminating irrelevant or redundant features that could introduce noise or lead to overfitting. By reducing the number of features, feature selection helps simplify models, improve generalization to unseen data, and enhance computational efficiency and interpretability [9] and [10]. This is a general quote and independent of the type of mapping applied to inputs. To further address this important problem, there are two related directions with different purposes: variable importance analysis and feature selection, which have some algorithms in common.

Variable importance analysis (VIA) is a set of techniques used to interpret machine learning models by assessing the contribution of each feature to the model's predictions [11] and [12]. Their primary goal is to understand which features significantly impact the model's output, thereby enhancing the model's interpretability. This is especially important in complex models, where the relationships between features and predictions may not be immediately clear. Generally, they are divided into model-free and model-based approaches [13] and [9]. Model-free methods utilize the inherent properties of data to determine the relevance scores. A simple approach to this category is correlation analysis measuring the strength of linearity between input and output variables [1], [13], [14], [15], and [16]. It deals with the computation of the correlation coefficient (CC) $r_i$ between input variable $x_i$ and target variable $y$, where it is assumed there is no dependency between input variables $x_i$. The CC $r_i$ takes values between -1 and 1, where 1 and -1 indicate strong positive and negative linearity between variables, respectively. The CC value of 0 states no linear dependency between variables [16]. This analysis can be translated to linear parametric modeling, where the target variable is expressed as a linear combination of input features. The weight of each input feature, standard regression coefficient (SRC), can be interpreted as the correlation between input features and target variable, provided that the unit of all input features is the same [13]. The correlation analysis does not apply to nonlinear dependent models in the present form. Ranked correlation coefficient (RCC) is an alternative to computing the dependency in nonlinear interaction models [17]. In the case of correlated variables, a partial correlation coefficient (PCC) is used. PCC performs the correlation analysis on uncorrelated parts of the input variable $x_i$ and target variable $y$ [18]. The statistical test approach is another approach to determine the dependency between variables [19]. The core idea of using statistical tests is to decide the dependency between variables by comparing specific statistics with a predefined criterion for rejecting or accepting the null hypothesis. Tests like common mean (CMN), common median, mutual information [20], [21], [22], and [23], statistical independence test, and squared ranked difference (SRD) [24] are of some common methods are used for measuring the dependency. Another important direction for variable importance analysis is to use non-parametric modeling. Locally weighted regression (LOESS) [25] is of this family aiming at finding a surrogate interpretable model, linear models, which compute the importance of each variable locally based on the regressed coefficient [13]. Generalized additive model (GAM) [26], pursuit projection regression (PPR) [27], and tree-based models are placed within this category. Tree-based models are heavily studied; the interested reader may be referred to [28] for further studies.

On the other hand, model-based methods require an analytical model. Permutation feature importance (PFI) [29, 13] is an approach employed to determine input variables' relevance score by permuting the feature of interest values and comparing the obtained prediction. However, it is limited to the case of uncorrelated features. Gradient-based methods are the more well-known and frequently used approach for determining the importance of the score of input variables. They compute the derivative of the target variable, or the loss function, concerning input variables or corresponding weights to measure how the input change contributes to the change in output. Sobol indices [30] is a well-known method for measuring variables' dependency. Gradient-based methods can also be seen from the perspective of variance-based decomposition methods [31], where the first-order variance approximation of the extended predictive function is used to infer the amount of dependency between variables. In this regard, the output variance is parametrized as a linear combination of feature variances, where the coefficient is squared of sensitivity of prediction target $y$ concerning input $x_i$. Consider the nonlinear mapping $y = f(x_1, ..., x_D)$, the first order Taylor expansion around the point $x_0$ is termed as:

$$y \approx f(x_0) + \sum_{i=1}^{D} \frac{\partial f}{\partial x_i}(x_i - x_{i0}) \tag{3}$$

The variance of the output is then computed as:

$$\text{var}[y] = \sum_{i=1}^{D} \left[\frac{\partial f}{\partial x_i}\right]^2 \text{var}[x_i] \tag{4}$$

Sometimes, the normalized form of relationship (2) is used for computing the relevance score of each feature as $\frac{\left[\frac{\partial f}{\partial x_i}\right]^2}{\text{var}[y]}$, indicating which portion of output uncertainty is explained by input uncertainty. The obtained relationship is valid for uncorrelated cases. The output variance in terms of dependency between variables can be extended as a relationship (3):

$$\text{var}[y] = \sum_{i=1}^{D} \left[\frac{\partial f}{\partial x_i}\right]^2 \text{var}[x_i] + \sum_{i=1}^{D} \sum_{j=1, i \neq j}^{D} \frac{\partial f}{\partial x_i} \cdot \frac{\partial f}{\partial x_j} \text{cov}[x_i, x_j] \tag{5}$$

It is worth mentioning that the obtained measure uses the local information to infer the importance of features. The global impact of each feature can be obtained by integrating the obtained indices over the input domain, leading to the Sobol index. Another important direction, an extension over the relationship (3), is to analyze the importance of features via variance decomposition methods [13]. The most well-known method within the former category is the analysis of variance (ANOVA) [32, 31], decomposing the output function into the summation of some terms, including the individual effects of each variable, 2-way interactions of variables, 3-way interaction of variables, and so on. To make a model interpretable, only 2-way interactions are considered in practice [26] since the number of interactions to test grows exponentially with interaction order.

As mentioned earlier, variable importance analysis (VIA) methods are primarily used for interpretability. This convention is shared across various fields in data-driven modeling. Similarly, feature selection serves the same purpose as VIA methods, helping to enhance interpretability by identifying relevant features. Feature selection methods are divided into three main categories: filter-based methods, aligned with the model-free approach of VIA methods; wrapper methods, aligned with the model-based approach of VIA techniques; and embedding techniques. The same set of methods can be used for each category and corresponding terminology used in VIA. An important technique not mentioned in VIA surveys is recursive feature elimination (RFE). RFE methods aim to remove the less relevant features from the feature sets recursively. RFE methods are founded on developing a surrogate model equipped with an explicit notion of scores like gradient scores, impurity index, variance-based measures, and so on [33]. The mentioned methods for computing each feature's importance scores are either post-processing or pre-identification approaches. Embedding techniques cover the gap between those via embedding feature selection mechanisms into the training process, saving time and computational sources for feature selection and predictive modeling. For the mentioned reason and to save the modeling effort, there is a penchant for using embedding techniques for feature selection. In this regard, [34] proposes a functional ANOVA using Variational auto-encoders to learn a decomposable latent space while training. Forward selection, backward elimination, stepwise learning, and LASSO methods [35, 25] are well-known embedding techniques utilizing the prediction power of input variables as their importance score while training. There are a wide range of methods that address the mentioned methods to which interested readers may be referred [36], [37], and [38].

In fact, the importance of feature selection in machine learning dates back decades, as it has always been a critical step in improving models' interpretability, efficiency, and performance. While the field of feature selection has evolved alongside advancements in machine learning, its significance has been amplified in the context of deep learning due to the increasing complexity and scale of these models. Deep learning models, particularly neural networks, have gained paramount attention recently because of their exceptional ability to model highly nonlinear and complex systems. These models, often over-parameterized and black-box systems, have revolutionized fields such as computer vision, natural language processing, and autonomous systems. Their power lies in their capacity to learn intricate patterns from vast amounts of data, often with millions of parameters, making them incredibly effective at handling tasks that involve large and unstructured datasets. However, this same capacity for complexity presents challenges, particularly in understanding and interpreting these models' decisions. Importance analysis in neural networks essentially concentrates on visualization to reveal where the model pays attention. Methods like saliency map [39], CAM [40], and Grad-CAM [41]. As a gate for interpreting deep neural networks, the feature selection part aims to obtain a sparse input layer and is a special case of sparse neural networks. Sparsity in neural networks is obtained via regularization or pruning. Applying regularization cannot essentially ensure the sparsity in the network at a single round of usage, and they need post-processing as well. Pruning is the most frequent method that is used to prune the intermediate layers to reduce the computational effort and required storage for performing inference [42], [43], and [44]. For instance, relevance score propagation is presented in [45], considering the change in output concerning the activation of each layer as relevance score. The same methodology can be applied to the input layer to select features. In [46], a method is introduced that uses a gated input architecture to score the importance of the feature via continuous relaxation of the Bernoulli distribution. The adopted approach considers a proxy variable $z$ parameterized using a binary variable to decide if an input variable is important. [47] introduces a drop-in layer, parameterized as element-wise production of input features and some weights. [48] introduces a model called LassoNet, which adds a residual connection between the input layer and the output to measure the importance of input variables via a direct link. DeepLasso is introduced in [49] to incorporate the gradient of output concerning input into the cost function. It is stated that the gradient of irrelevant variables will approach zero. NeuroFS is introduced in [50] to improve feature selection by gradually pruning uninformative features from the input layer, resulting in an efficient and informative subset of features. In [51], gradient ensemble feature selection inspired by a sparse neural network is proposed to measure the neuron importance as the sensitivity of the loss to the input variables.

As mentioned earlier, the existing approaches, especially those dealing with correlated variables, can be used to identify the importance score in the context of nonlinear system identification. However, some works are specifically relevant to this context of interest. The study in [3] is of earlier work that presents a model-free variable selection. The unknown function, which should be estimated based on available data, is assumed to be smooth so that the largest Lipchitz quotient can be computed from the data as a bound. Removing those variables that contribute more to the prediction will force the Lipchitz quotient to exceed the calculated bound. In [4], another model-free variable selection is given, working based on an estimation of goodness of fit. It is stated that such criteria as global sensitivity analysis, mutual information, and distance correlation for variable selection can be misleading, and goodness of fit is an appropriate criterion for selecting variables. In [52], a variable selection procedure is given based on the derivative squared average estimation. There is no restriction on the unknown function, so the function and its derivative are unknown. Two algorithms are given to approximate the function and its partial derivative.

In all the aforementioned methods, the proposed approach functions as an end-to-end process where importance scores are learned to optimize predictive performance. This is achieved indirectly via sparsification of the network, meaning there is no explicit mechanism guiding the feature selection process based on the definition of importance scores, which typically quantify the contribution of input variables (or sets of variables) to the model's output rather than directly addressing uncertainty. To create more control over the learning curve of importance score, we introduce an end-to-end training methodology, an embedded feature selection technique, for black-box models equipped with a guidance term in its objective function to output a meaningful importance score. The adopted approach uses the Fisher Information Matrix (FIM) to encode the importance score of input variables. The FIM provides a way to quantify the information content of an observable random variable concerning unknown parameters within a model that characterizes the variable. When parameters in a model are directly linked to individual features, the diagonal elements of the FIM can signify the relative importance of each feature. We introduce a proxy variable $\alpha_{t-i}$ indicating the importance score of dynamic input $x_{t-i}$ where the output target is parameterized as $f_\theta(x_{t-\tau}\odot\alpha_{t-\tau}, \ldots, x_t\odot\alpha_t)$. Alternatively, the importance scores $\alpha_{t-i}$ can be seen as a modulating factor applied to columns of the first layer weight matrix. By doing so, it can be easily shown that the importance of each column of weight matrix, or rather the input neurons, is inversely related to the FIM, which is computationally infeasible (for large dynamic systems of length $T$, it is of order $O(T^3)$). One can see the importance of score as a nonlinear function of FIM to solve this issue. With this in mind, we propose a nonlinear function decision unit to take the FIM as input and output the score. To encode the relative importance, the output must be scaled within 0 and 1 to be seen as a binary variable regarding whether a time step is important. As no information is related to important time steps, we add a variance regularization term to enforce the decision unit to generate importance scores that mostly contribute to the variance output. A similar work to ours is [53], which utilizes FIM to prune a neural network's intermediate neurons. The proposed approach uses diagonal elements of FIM as an importance score. The larger the elements of FIM are, the more important the corresponding neurons are. However, the introduced approach prunes neurons without considering their interaction. There may be cases where the neurons contribute to the prediction in the presence of other neurons, meaning the full matrix of FIM should be considered rather than its diagonal elements. The results obtained in Table 1 show that the adopted approach is superior to existing methods in capturing the interaction between features.

**Notations**: The vectors are shown by bold lower case letter $\boldsymbol{x}$, $\boldsymbol{a^l}$, $\boldsymbol{b^l}$, and the matrices by bold uppercase letter $\boldsymbol{W^l}$, except for $T$ indicating the total sequence length, while scaler with unbold lower case letter x. $\boldsymbol{a^m}$, $\boldsymbol{W^m}$ and $\boldsymbol{b^m}$ indicate the activation representation, weight matrix, and bias vector associated with layer $m$, respectively. Activation functions are denoted by $\phi$.

$J_{W^{(m)}}l(y,\hat{y};\theta)$ and $H_{W^{(m)}}l(y,\hat{y};\theta)$ denotes the Jacobian and Hessian matrix of loss function $l(y,\hat{y};\theta)$ for the parameters $W^{(m)}$ of layer $m$ respectively. Furthermore, $\theta$ is the set of all parameters required to be adjust $\theta = \{\theta_f, \theta_D\}$ where $\theta_f = \{(W^{(1)}, b^{(1)}), \ldots, (W^{(m)}, b^{(m)})\}$ and $\theta_D$ is the parameter of the decision unit.

## 2. Problem Setup and Background

Let $u_t \subset R^{D_u} = \{u_1^t, \ldots, u_{D_u}^t\}$ be the input vector and $y_t \subset R^{D_y} = \{y_1^t, \ldots, y_{D_y}^t\}$ the output vector of the system, where extracted from the probability distribution $P_{u,y}$ over time steps $t = 1, \ldots, T$. Define $x_i \subset R^{(D_u+D_y)\tau} = \{u_1^{i+1:i+\tau-1}, \ldots u_{D_u}^{i+1:i+\tau-1}, y_1^{i+1:i+\tau-1}, \ldots y_{D_y}^{i+1:i+\tau-1}\}$, $i = 0, \ldots, t-\tau$ as input model with corresponding target vector $y_i \subset R = y^{i+\tau}$. To make the expression simple, we consider a single input single output case, $D_u = D_y = 1$, where generalization to multivariate cases is straightforward. Now, the problem of black-box nonlinear system identification is formulated as $\hat{y}_i = f_\theta(x_i)$ where $f_\theta$ is parameterized using a neural network. Give a loss function $l(y_i, \hat{y}_i; \theta_f)$, the parameters $\theta_f$ are obtained via minimizing the loss function $l(y_i, f_{\theta_f}(x_i))$ using gradient descent optimization techniques. By defining a set of proxy variable $\alpha_i \subset R^{2\tau} = \{\alpha_1, \ldots, \alpha_{2\tau}\}$ where $\alpha_i \in \{0,1\}^{2\tau}$ (the dimension is valid for the SISO case, yet in general, the dimension should be aligned with $(D_u + D_y)\tau$ ) indicating the importance of dynamics involved in $x_i$, the loss function is now changed to $l(y_i, f_{\theta_f}(x_i \odot \alpha_i))$ where $\odot$ is element-wise production. For a neural network with activation function $\phi$, the first layer activation is now computed as the following:

$$a^{(1)} = \phi(W^{(1)}x_i \odot \alpha_i + b^{(1)}) = \phi\left[\sum_{j=1}^{2\tau} W^1[:,j]x_j \odot \alpha_j\right] \quad (6)$$

where $W[:,j]$ is the jth column of the weight matrix $W^{(1)}$. The relationship (6) uses the simple fact in matrix multiplication by a vector, the vector elements (take $x_j \odot \alpha_j$) is linearly combined with the matrix columns. It means the factor modulates each input neuron's weights $\alpha_j$. In a better word, the relevance score of each system dynamics can be seen as the importance of each column of the weight matrix. This view collides with automatic relevance determination for feature selection learned by the model. In [46], the proxy variables $\alpha_i$ is considered as weights of a layer, a so-called drop-in layer whose output is computed as $x_j \odot \alpha_j$) Therefore, one can treat the proxy variables as parameters and obtain them via gradient descent optimization (7).

$$\theta_f^*, \alpha^* = \arg\min_{\theta_f, \alpha} l(y_i, f_{\theta_f}(x_i \odot \alpha_i); \theta_f) \quad (7)$$

The scheme for the associated approach is shown in Figure 1 [46]. This approach's main drawback is that variables' importance scores are determined independently of the correlations between different time steps, overlooking the dynamic relationships between input variables.

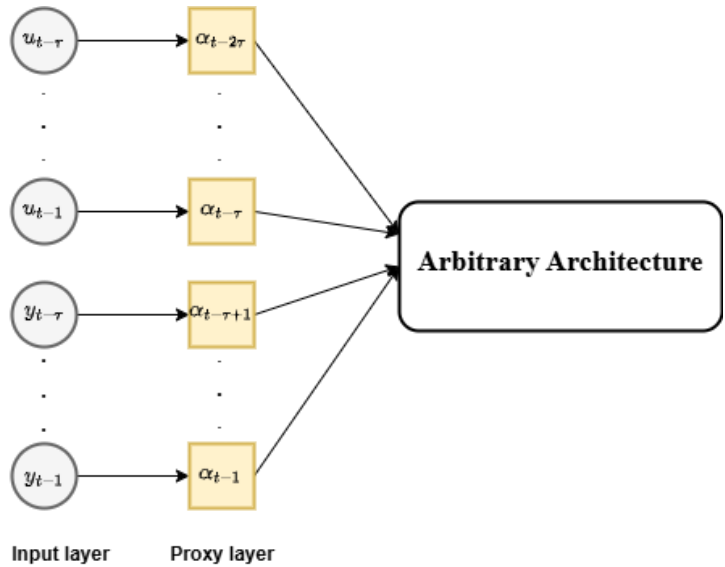

**Figure 1** – Drop-in layer approach for variable importance determination [46]

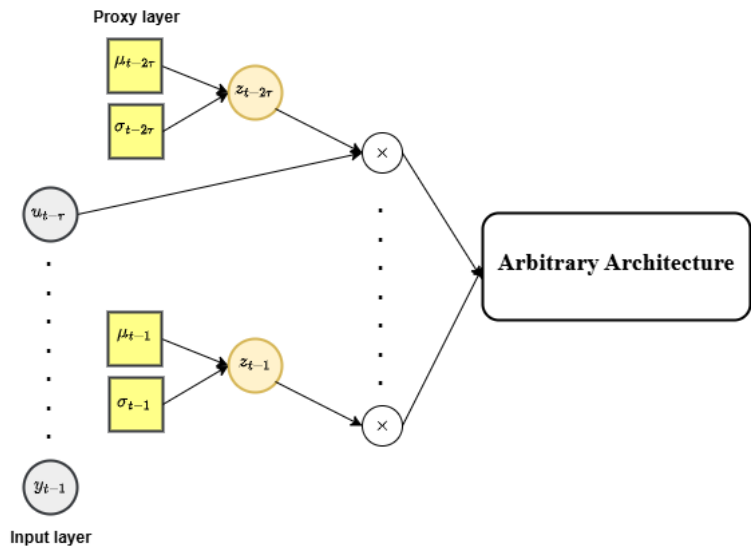

**Figure 2** – Stochastic gated input for variable importance determination [47]

A similar approach involves treating the importance score as binary variables and introducing randomness into the selection process, where the probability of a time step being active is determined and learned during the network's training [47]. The proposed network can be viewed as a continuous relaxation of the $l_0$-norm used in the feature selection problem. The corresponding scheme for this approach is shown in Figure 2, Where $z_i$ is defined as $\max(0, \min(1, \mu_i + \sigma_i))$. Like the drop-in layer approach, the parameters $\mu$ and $\sigma$ can be adjusted through stochastic gradient optimization. The stochastic gated input is anticipated to encounter issues similar to correlated features. While the training process does not directly reveal the interactions between correlated features (it is evident from the way the variances are parameterized), it is claimed that the proposed method can capture such interactions. Because of the resemblance between these approaches and ours, we consider them as baselines for comparing our results. The adopted approach is the so-called stochastic gated input.

## 3. Proposed Method

As already mentioned, the proxy variables can be seen as an important measure of the column weights of the first layer of the network, so the importance scores of input dynamics can be seen as the importance of each column vector.

To begin with, let's start with the objective function that is going to be optimized, in general, expressed as $E_{P_{x,y}}[\log p(y|x;\theta)]$ where $\theta$ is the parameter set of the model. The regression task is reduced to empirical approximation $\frac{1}{N}\epsilon(\theta_f)^T\epsilon(\theta_f)$, where $\epsilon(\theta_f) = y - \hat{y}$ and $\epsilon_i(\theta_f) = y_i - \hat{y}_i$ is an element of $\epsilon(\theta_f)$. In addition, the FIM is empirically approximated by:

$$E\left[\frac{\partial \log p(y|x;\theta_f)}{\partial \theta_f}\left(\frac{\partial \log p(y|x;\theta_f)}{\partial \theta_f}\right)^T\right] = \frac{1}{N}J_{\theta_f}\epsilon(\theta_f)^T J_{\theta_f}\epsilon(\theta_f) \tag{8}$$

The relationship (8) encodes the variance of the sensitivity of the error function concerning the parameter $\theta_f$, defining the precision of the weight parameters in a weight-feature correspondence model. Consider the weight matrix of the first layer as $W^{(1)} \in R^{h \times 2\tau}$, then $FIM(W^{(1)}) \in R^{2h\tau \times 2h\tau}$, meaning the computation of the FIM matrix linearly increases as the input dynamic length increases. To compute the relevance score associated with each input dynamics, consider the Newton method for updating the weight matrix:

$$W^{(1)}_{k+1} = W^{(1)}_k - \eta\left[H_{W^{(1)}}l(y,\hat{y};\theta_f)\right]^{-1}[J_{W^{(1)}}l(y,\hat{y};\theta_f)]$$

where $\eta$ is the learning rate, $J_{W^{(1)}}$ is the Jacobian matrix of the loss function concerning $W^{(1)}$, and $H_{W^{(1)}}$ is the Hessian matrix. The local covariance of the first layer weight matrices over a mini-batch of samples is computed as the following:

$$\text{cov}\left(W^{(1)}_{k+1} - W^{(1)}_k\right) = \text{cov}\left([H_{W^{(1)}}l]^{-1}[J_{W^{(1)}}l]\right) \tag{9}$$

where $l(y,\hat{y};\Theta) = \frac{1}{N}\epsilon(\theta)^T\epsilon(\theta)$. So, by definition, we will have:

$$\nabla_{\theta^{(1)}}l(y,\hat{y};\Theta) = \frac{1}{N}J_{W^{(1)}}\epsilon(\theta_f)^T\epsilon(\theta_f) \tag{10}$$

$$\text{cov}\left([H_{W^{(1)}}l]^{-1}\frac{1}{N}J_{W^{(1)}}\epsilon(\theta_f)^T\epsilon(\theta_f)\right)$$
$$= [H_{W^{(1)}}l]^{-1}\frac{1}{N}J_{W^{(1)}}\epsilon(\theta_f)^T J_{W^{(1)}}\epsilon(\theta_f)[H_{W^{(1)}}l]^{-1}\text{cov}\left(\epsilon(\theta_f)\right) \tag{11}$$

Where under mild conditions, the $\frac{1}{N}J_{W^{(1)}}\epsilon(\theta_f)^T J_{W^{(1)}}\epsilon(\theta_f)$, or rather FIM as shown in relationship (8), equals to $H_{W^{(1)}}l$. So we will have:

$$\text{cov}\left([H_{W^{(1)}}l]^{-1}\frac{1}{N}\nabla_{W^{(1)}}\epsilon(\theta_f)^T\epsilon(\theta_f)\right) = [\text{FIM}(W^{(1)})]^{-1}\sigma^2 \tag{12}$$

where $\sigma^2$ is the variance of error. The derived relationship is known as the Cramer-Rao lower bound. When the input features are independent, the precision of the estimated parameters is proportional to the inverse of the diagonal elements of the Fisher Information Matrix (FIM). Parameters corresponding to directions with large diagonal elements of the FIM are estimated with lower precision and vice versa. However, this assumption does not hold in our case, as it relies on the independence of input features, which is not valid for dynamic systems. A potential solution is to use the inverse of the full FIM; however, it is computationally infeasible for large dynamic systems due to its complexity, which scales as $O(\tau^3 h^3)$, where $\tau$ represents the length of input dynamics. It indicates that the computation of FIM for the first layer is cubically scaled concerning the network parameter, making it infeasible for a larger network. To mediate the problem, we need to expand the $\text{FIM}(W^{(1)})$. In neural networks, calculating the FIM typically involves using the Jacobian of the loss function for the first layer parameters for the loss function, as shown in (8). This process is straightforward, as the Jacobian for the input layer is computed as:

$$J_{W^{(1)}}\epsilon(\theta_f) = \delta^{(1)}.(x^{(1)})^T \tag{13}$$

where $J_{W^{(1)}} \in R^{1 \times 2h\tau}$ and $\delta^{(1)}$ is the back-propagated error from the next layer to the loss function and $x^l$ is the input vector to layer 1. Then FIM would be:

$$\text{FIM}(W^{(1)}) = \frac{1}{N}J_{W^{(1)}}\epsilon(\theta_f)^T J_{W^{(1)}}\epsilon(\theta) = \frac{1}{N}x^{(1)}\delta^{(1)^T}\delta^{(1)}.(x^{(1)})^T \propto \frac{1}{N}x^{(1)}(x^{(1)})^T$$

The above relationship denotes that the FIM is related to the input covariance matrix for zero mean input, modulated by some sensitivity terms. So, we can say that FIM is affected by the correlation matrix of inputs as the following:

$$FIM(W^{(1)}) \propto C_x \tag{14}$$

where $C_x$ is computed as (15):

$$C_x = \begin{pmatrix} E(u(t-\tau)u(t-\tau)^T) & \cdots & E(u(t-\tau)y(t-1)^T) \\ \vdots & \ddots & \vdots \\ E(y(t-1)u(t-\tau)^T) & \cdots & E(y(t-1)y(t-1)^T) \end{pmatrix}$$
$$= \begin{pmatrix} C_u(0) & \cdots & C_{ut}(\tau-1) \\ \vdots & \ddots & \vdots \\ C_{ut}(\tau-1) & \cdots & C_y(0) \end{pmatrix} \in R^{2\tau \times 2\tau} \tag{15}$$

where the entities in (15) is defined as $C_u(t_1 - t_2) = E(u(t_1)u(t_2)^T)$, $C_y(t_1 - t_2) = E(y(t_1)y(t_2)^T)$, and $C_{uy}(t_1 - t_2) = E(u(t_1)y(t_2)^T)$ are called auto-covariance of input, output-covariance, and cross-covariance between input and output, respectively.

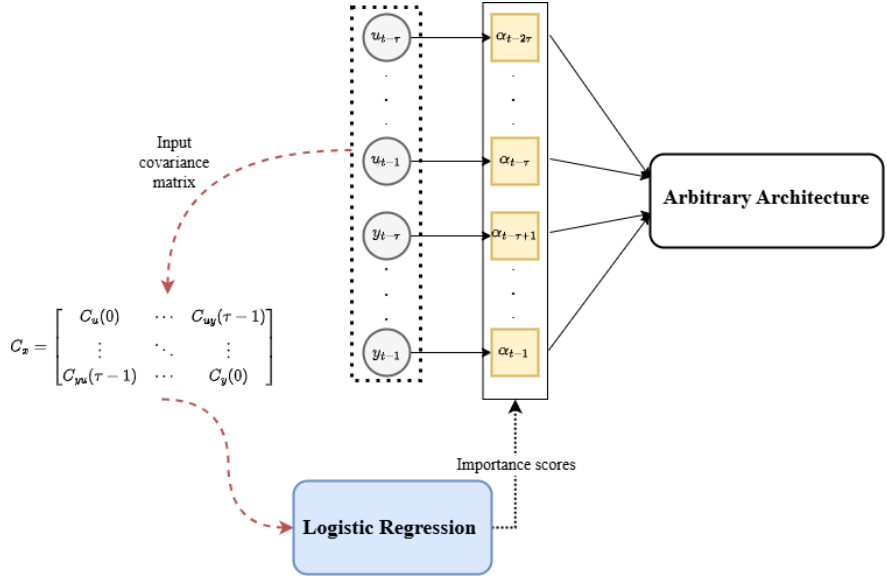

**Figure 3** – The proposed approach for dynamic selection. The full FIM feeds the decision unit logistic regression and outputs the relevance score of each time step.

The time complexity of computing the correlation matrix $C_x$ is $O(\tau^2)$, which is independent of the network parameters. While the proxy methods used to estimate the importance scores, as outlined in equation (14), can help mitigate the computational burden of calculating the Fisher Information Matrix (FIM), they are not suitable for assessing the dynamics of the model for two key reasons. First, the issue of efficiently calculating the inverse of the FIM remains unresolved. Second, the approximations used to derive the importance scores result in the loss of crucial information related to the model's predictive power, particularly the uncertainty of the model's output. To address the first issue, we approach the importance scores as a nonlinear function of the FIM with a parameter set $\boldsymbol{\theta_D}$, extending a simple nonlinear model that takes the FIM as input and outputs the importance scores $\boldsymbol{\alpha_i}$. To scale the importance scores between 0 and 1, we apply a sigmoid activation function at the output layer. The architecture of the proposed model is shown in Figure 3. The logistic regression model serves as a decision unit, taking the approximation of FIM, as obtained in (14), as input and producing importance scores for each dynamic. This approach makes sense as we have observed from (12) that the precision of weight parameters is a nonlinear function of FIM. To solve the second problem, we must incorporate the uncertainty information, lost in FIM approximation, into the training process to guide the network in selecting dynamics that mostly contribute to the output uncertainty.

To solve the issues, consider the first-order expansion of the predictive function around $x_0$.

$$\boldsymbol{y_i} = f_\theta(\boldsymbol{x_i} \odot \boldsymbol{\alpha_i}) = f_\theta(\boldsymbol{x_0} \odot \boldsymbol{\alpha_i}) + \nabla_\mathrm{x} f_\theta(\boldsymbol{x_0} \odot \boldsymbol{\alpha_i})[(\boldsymbol{x_i} - \boldsymbol{x_0}) \odot \boldsymbol{\alpha_i}]$$

The first order variance approximation then would be:

$$\mathrm{var}[\boldsymbol{y_i}] = \boldsymbol{g}^T \mathbf{cov}(\Delta \boldsymbol{x_i} \odot \boldsymbol{\alpha_i}) \boldsymbol{g} = \boldsymbol{g}^T \mathbf{diag}\{\boldsymbol{\alpha_i}\} \boldsymbol{C_x} \mathbf{diag}\{\boldsymbol{\alpha_i}\} \boldsymbol{g} \tag{16}$$

where $g = \nabla_\mathrm{x} f_{\boldsymbol{\theta_f}}(\boldsymbol{x_0} \odot \boldsymbol{\alpha_i})$ and $\boldsymbol{C_x}$ is computed as (15). To regulate the values of $\boldsymbol{\alpha_i}$ to be aligned with the output uncertainty, we add the obtained variance regularization term to the cost function to guide the decision unit module to produce $\boldsymbol{\alpha_i}$ which is mostly aligned with output uncertainty. Therefore, the total object function would be as follows:

$$\boldsymbol{\theta}^* = \arg\min_{\theta} \left\{ \frac{1}{T} \left( \sum_{i=1}^{T}(\boldsymbol{y_i} - \hat{\boldsymbol{y}}_i)^2 + \lambda_v \big(\text{var}[y_i] - \boldsymbol{g}^T \text{diag}\{\boldsymbol{\alpha_i}\} \boldsymbol{C_x} \text{diag}\{\boldsymbol{\alpha_i}\} \boldsymbol{g}\big)^2 \right) \right\} \tag{17}$$

where $\boldsymbol{\theta}$ includes the parameters of the decision unit $\boldsymbol{\theta_D}$ and the model itself $\boldsymbol{\theta_f}$.

## 4. Simulation Results

In this section, we discuss our experiments on both simulated and real-world systems to show the performance of our study on feature importance detection. We consider some fictional nonlinear systems with different complexity to examine the presented work. The results are compared to the schema presented in Figures 1 and 2, called drop-in layer and stochastic gated input, respectively. The only difference between the proposed method and the mentioned works relies on how they compute the importance score; the adopted architecture for all three is considered the same. Results show that proposed approach can capture any complex interaction among input variables.

### 4.1. Test Suits of Fictional Nonlinear Systems

The list of simulated systems is mentioned in Table 1, covering systems with different nonlinearities and structures.

**Table 1** – Test suites of data generation functions

| | |
|---|---|
| $F_1$ | $\sin(y_{t-1}) + 0.01 y_{t-2} + u_{t-4} + u_{t-1}^2 + u_{t-2} u_{t-3}$ |
| $F_2$ | $0.01 y_{t-1}^2 + u_{t-1}^5 + u_{t-2} u_{t-3} u_{t-4}^4$ |
| $F_3$ | $u_{t-1}^2 + u_{t-2} u_{t-3} u_{t-4}$ |
| $F_4$ | $u_{t-3} u_{t-2} + u_{t-3} u_{t-1} + u_{t-3} u_{t-2} u_{t-1} + \sin(y_{t-2}) + \exp(-y_{t-1})$ |
| $F_5$ | $\sin(u_{t-1} u_{t-2}) + \exp(-y_{t-1} y_{t-2})$ |
| $F_6$ | $\exp(\sin y_{t-1}) + y_{t-3} \exp(-u_{t-2})$ |
| $F_7$ | $u_{t-5} \exp(\sin y_{t-1}) + y_{t-3} \exp(-u_{t-2})$ |
| $F_8$ | $\exp(|u_{t-1} + u_{t-3}|) + u_{t-2} u_{t-4} + \dfrac{1}{1 + y_{t-6}^2}$ |
| $F_9$ | $\sqrt{\exp(u_{t-5})} + \dfrac{1}{1 + y_{t-6}^2 + u_{t-2}^2}$ |
| $F_{10}$ | $2^{-|u_{t-1} u_{t-2}|} \sqrt{y_{t-1}} + 0.01 \arcsin(y_{t-10})$ |
| $F_{11}$ | $0.01\, u_{t-10} \arctan(y_{t-1} + u_{t-1}) + \max(u_{t-2}, 0.5) + \dfrac{1}{1 + y_{t-5}^2 + u_{t-3}^2}$ |

For each simulated system, we generate 16,000 samples for training and 10,000 samples for testing, using a uniformly distributed input signal $u$ in the range [-2.5, 2.5]. The results presented are obtained by averaging over 100 identified models for each simulated system, ensuring the robustness and reliability of the analysis. A time-delayed neural network is trained with a consistent architecture across all models to approximate the specified functions. This network considers a lag of 10 for both the input and output of the systems, where the feature vector is defined as $x_t = [u_{t-10}, ..., u_{t-1}, y_{t-10}, ..., y_{t-1}]$.

To optimize space, the results for four selected simulated systems are shown in Figure 4 (a-d), where the importance scores of system dynamics are visualized as bar plots. These bar plots effectively illustrate how the importance of different system dynamics varies across the simulated systems. Comprehensive results for all listed systems, including detailed comparisons, are provided in Table 2. It is important to note that in the stochastic gated input approach, only the mean of the Gaussian distribution needs to be trained, while the variance is fixed at 1. This design choice simplifies the training process and focuses the model's capacity on learning the most relevant features of the system dynamics. As illustrated, the predictive performance of all three models is nearly identical, highlighting their comparable ability to predict outcomes accurately. However, their effectiveness in capturing the importance dynamics differs significantly, as depicted in the accompanying bar charts. The proposed method demonstrates a clear advantage in inferring these dynamics among the three. This superiority is attributed to its use of a sigmoid layer, allowing it to encode each time step's relative importance effectively.

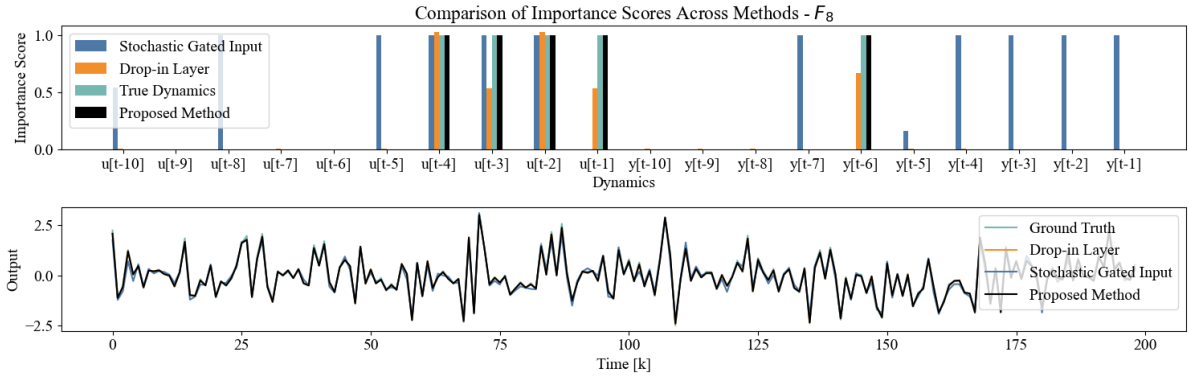

**(a)** – Results for nonlinear test suit $F_8$

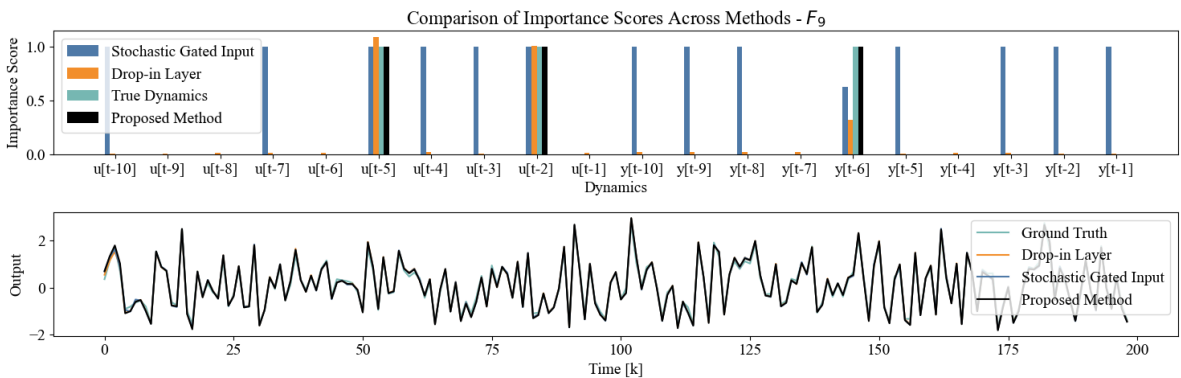

**(b)** – Results for nonlinear test suit $F_9$

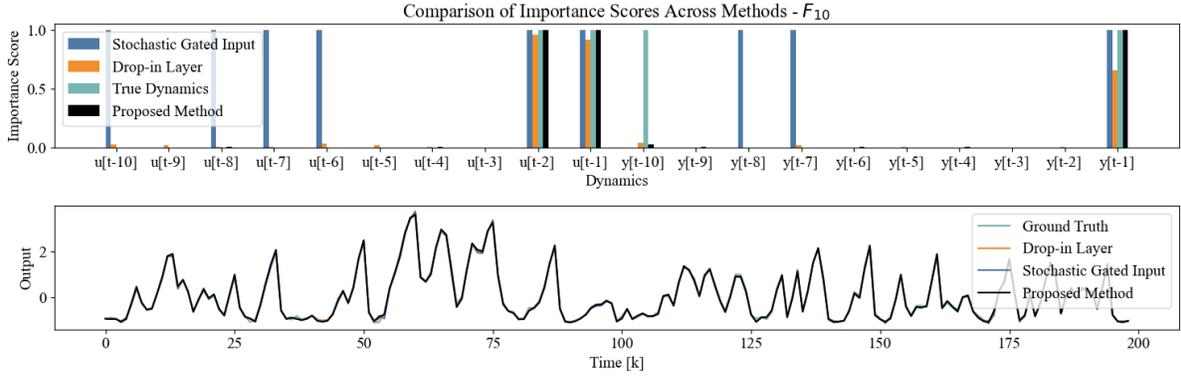

**(c)** – Results for nonlinear test suit $F_{10}$

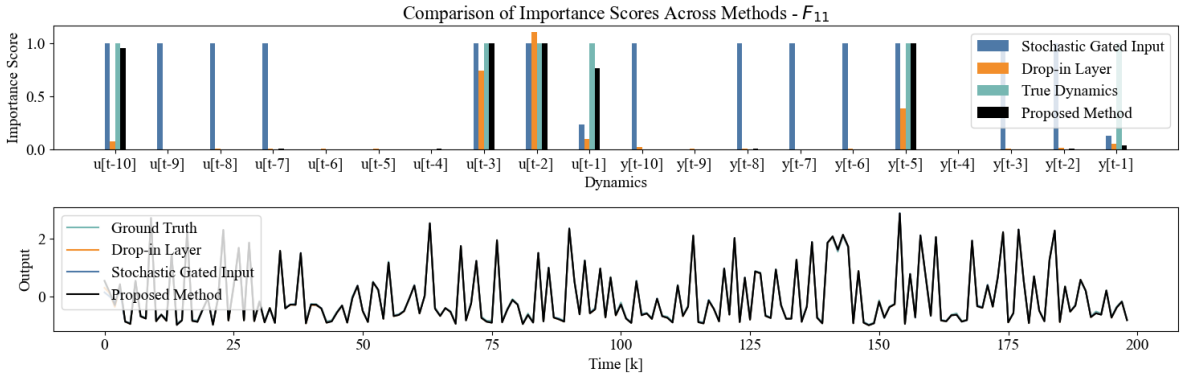

**(d)** Results for nonlinear test suit $F_{11}$

Figure 4 – The result of simulated systems $F_8, F_9, F_{10}$ and $F_{11}$

The drop-in layer method also shows potential in detecting important dynamics. Still, its relative importance scores are less comparable, making it less reliable for evaluating the relative contributions of different time steps. Additionally, while the drop-in layer introduces a certain sparsity level, it is not as pronounced as observed in the stochastic gated input method or our approach. For instance, consider the experimental results for $F_8$ The magnitude of scores computed by the drop-in layer method does not align well with the underlying dynamics. In contrast, the proposed method effectively captures the dynamics with greater accuracy. Overall, the performance of the drop-in layer surpasses that of the stochastic gated input method in identifying system dynamics, but it still falls short compared to our approach.

To comprehensively compare all aspects of the experimental results, we calculated the L1-norm of the importance scores as a sparsity measure, as summarized in Table 1. These results further highlight the effectiveness of the proposed method in capturing and representing importance dynamics more accurately and consistently than the alternative approaches. We ensure all three models share the same architecture and parameter count during inference for a fair comparison. However, our method includes an additional decision unit during training, increasing its parameter size. This unit computes importance scores to assess input dynamics but is excluded during validation and inference. The calculated scores are instead used to weigh the input dynamics for improved output prediction. This design balances efficient inference with enhanced training performance.

As shown in Figure 4, the output of the decision unit for many cases can be interpreted as realizations of a random variable, often explicitly producing binary outputs of 0 or 1. This behavior highlights the probabilistic nature of the decision process, where the relevance scores represent the likelihood of input neurons contributing to the output variance. However, thresholding these relevance scores at 0.5 is important in achieving greater sparsity by selectively retaining only those variables whose contributions surpass the threshold of random significance. The choice of a 0.5 threshold is grounded in the probabilistic interpretation of relevance scores. Focusing on variables with scores greater than 0.5 ensures that the selected features exhibit stronger-than-random contributions to the output variance. This straightforward and consistent criterion aligns with the binary decision-making framework, offering a principled way to distinguish between essential and non-essential features in various systems. Thresholding becomes a valuable tool for simplifying the feature set and enhancing the interpretability of the model's decisions, especially when analyzing complex nonlinear systems.

It is important to note, however, that while we propose thresholding to improve sparsity, the relevance scores reported in this work have not been thresholded. This decision was intentional, allowing for a more comprehensive evaluation of the feature contributions without imposing sparsity constraints. For instance, the computed $L_1$-norms for all cases were derived from the full spectrum of relevance scores, providing a holistic view of the features' influence on system outputs. Thresholding, when applied, primarily serves to refine the feature set by eliminating less relevant inputs, thereby streamlining the analysis and interpretation.

Table 2 – Obtained result for drop-in layer [], stochastic gated input [], and the proposed method

| System | Drop-in Layer | | Stochastic Gated Input | | Proposed Model | |
|---|---|---|---|---|---|---|
| | MSE | $|\alpha|_1$ | MSE | $|\alpha|_1$ | MSE | $|\alpha|_1$ |
| $F_1$ | 0.0001 | 7.5 | 0.0001 | 7 | 0.0001 | 5.9 |
| $F_2$ | 0.0005 | 6.9 | 0.0006 | 9.7 | 0.0005 | 4.8 |
| $F_3$ | 0.0008 | 4 | 0.0007 | 6 | 0.0001 | 4 |
| $F_4$ | 0.001 | 7 | 0.001 | 13.1 | 0.001 | 4.95 |
| $F_5$ | 0.001 | 4.8 | 0.001 | 6.9 | 0.001 | 4.2 |
| $F_6$ | 0.001 | 4.1 | 0.001 | 5.5 | 0.002 | 3 |
| $F_7$ | 0.001 | 3 | 0.002 | 8.1 | 0.001 | 3.95 |
| $F_8$ | 0.007 | 13.1 | 0.007 | 2.5 | 0.007 | 4.8 |
| $F_9$ | 0.005 | 2.6 | 0.006 | 13.5 | 0.006 | 3.02 |
| $F_{10}$ | 0.005 | 3.9 | 0.004 | 9.3 | 0.005 | 3.1 |
| $F_{11}$ | 0.004 | 3.9 | 0.004 | 9.3 | 0.003 | 4.8 |

## 4-2- pH Neutralization Process

A laboratory experiment is conducted to demonstrate the efficacy of the proposed method. The inputs to the process include the flow rates of acid, water, and base, while the outputs are the fluid's pH value and the level of the mixing tank. The base flow rate acts as the controlled variable in the pH control loop, while the acid flow rate serves as a disturbance input. The level setpoint for the mixing tank is held constant for simplicity, and the level is regulated by adjusting the water flow rate. Given that the settling time of the level process can be disregarded, the influence of the level control loop on the pH control loop is minimal, allowing for the decoupling of the level control loop [54]. The flow rate of base and pH values are shown in Figure 5.

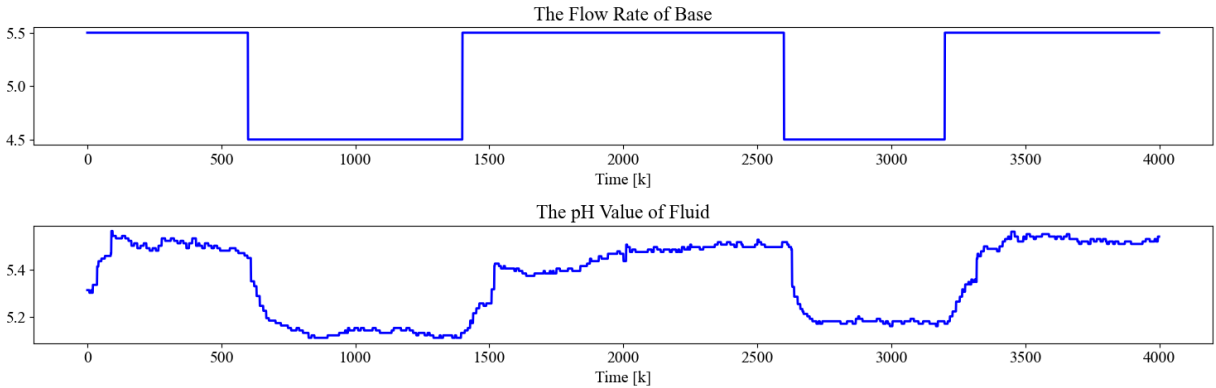

Figure 5 – The flow rate of base and pH value of fluid

Given the base and acid flow rate, we aim to predict the fluid's pH value. Since the acid flow rate is constant, the identification process is done based on the fluid's pH value and the base's flow rate [54]. For this purpose, we collected 4000 data, as illustrated in Figure 6. We chose 2000 sample points for training and 2000 for testing. We adopt the ARX model with $n_a = n_b = 5$ for all three models. The corresponding importance scores of each input dynamic are computed in Figure 6. Based on the calculated MSE values—0.001 for the proposed method, 0.003 for the stochastic gated input model, and 0.008 for the drop-in layer model—the proposed approach demonstrates superior performance, delivering better MSE results with a sparse input space than the other methods.

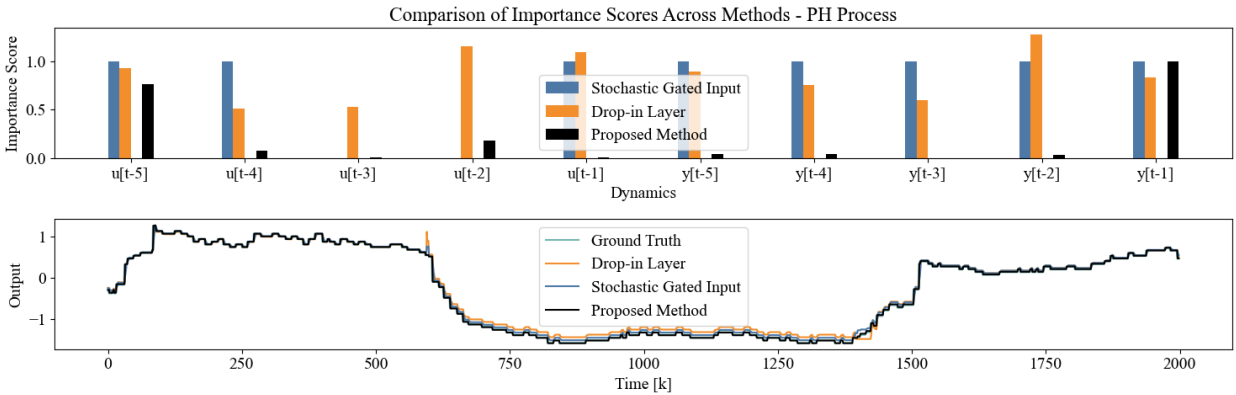

Figure 6 – The relevance score of input dynamics for the pH neutralization process

## 5. Conclusion

This paper presents an innovative framework for nonlinear system identification, combining relevant variable selection with nonlinear function approximation in a unified approach. A key feature is using a decision module based on logistic regression to compute relevance scores from the Fisher Information Matrix (FIM). These scores are then used to dynamically prune features, improving both model efficiency and interpretability. Importantly, the approach eliminates any additional computational overhead during inference, as the decision unit is not required once the scores are computed. By utilizing numerical simulations and a practical case study of an industrial pH neutralization process, we showcase the method's ability to reveal complex feature interactions while ensuring computational efficiency. The approach is particularly well-suited for real-time applications, as it focuses on the most relevant features without requiring recalculation of the FIM during inference.

**Future Works –** Future Research could explore integrating Graph Neural Networks (GNNs) [55] to model complex interdependencies between features more effectively, enabling improved handling of high-dimensional and dynamic data. The flexibility of GNNs could enhance the model's ability to capture relationships that are not purely linear, making it more applicable in various domains, such as control systems, robotics, and sensor networks. Additionally, transformers [56] —known for their capabilities in modeling long-range dependencies and input importance—could be utilized to capture temporal or sequential feature interactions. By leveraging the attention mechanism in transformers, the framework could further enhance the identification of critical features across time-series data, improving model robustness and interpretability. Another direction would be incorporating techniques from deep reinforcement learning [57] may offer new avenues to adjust relevance scores in real-time adaptively, improving model performance in dynamic environments.